# Evaluation of Causal Reasoning for Large Language Models in Contextualized Clinical Scenarios of Laboratory Test Interpretation

Balu Bhasuran[1], Mattia Prosperi[2], Karim Hanna[3], John Petrilli[3], Caretia JeLayne Washington[2], Zhe He[1, *]

[1]School of Information, Florida State University, Tallahassee, FL, USA; [2]Department of Epidemiology, University of Florida, Gainesville, FL, USA; [3]Morsani College of Medicine, University of South Florida, Tampa, FL, USA

**Abstract**

This study evaluates causal reasoning in large language models (LLMs) using 99 clinically grounded laboratory test scenarios mapped to Pearl's Ladder of Causation: association, intervention, and counterfactual reasoning. We focused on common lab tests such as Hemoglobin A1c (HbA1c), creatinine, and vitamin D, and paired them with clinically relevant causal factors, including age, gender, obesity, and smoking. Two LLMs GPT-o1 and Llama-3.2-8b-instruct were tested, with responses rated by four medically trained human experts. GPT-o1 demonstrated superior discriminative performance (AUROC overall = 0.80 ± 0.12) compared to Llama-3.2-8b-instruct (0.73 ± 0.15), with higher association (0.75 vs. 0.72), intervention (0.84 vs. 0.70), and counterfactual scores (0.84 vs. 0.69). Sensitivity (0.90 vs. 0.84) and specificity (0.93 vs. 0.80) were also greater for GPT-o1. Reasoning ratings followed similar trends. Both models performed best on intervention questions and worst on counterfactuals, particularly "altered outcome" scenarios. Findings suggest GPT-o1 offers more consistent causal reasoning, but further refinement is needed before high-stakes clinical deployment.

**Keywords:** causal reasoning; large language models; laboratory tests; clinical decision support; Pearl's Ladder of Causation

**Introduction**

The field of large language models (LLMs) has witnessed rapid and transformative advancements in recent years. Cutting-edge models such as ChatGPT-4[1] and Gemini 2.5 Pro[2] now demonstrate capabilities that emulate human reasoning and problem-solving across a wide range of professional domains. In medicine, these models often outperform clinicians on standardized benchmarks. For example, Med-PaLM 2[3]—a fine-tuned version of Google's PaLM (Pathways Language Model) model trained on medical data—achieves state-of-the-art performance with scores of 86.5% on MedQA (USMLE), 81.8% on PubMedQA, and 95.2% on Massive Multitask Language Understanding (MMLU) Professional Medicine. Its ensemble-refined configurations also lead in MMLU Clinical Knowledge (88.7%) and College Medicine (83.2%). Building on these benchmarks, researchers and clinicians have increasingly applied LLMs to real-world clinical tasks such as differential diagnosis prediction[4], discharge summary generation[5], patient portal message drafting[6], lab test interpretation[7], clinical note summarization[8], and clinical trial matching[9].

More recently, the focus of LLM research in healthcare has begun to shift from output accuracy to reasoning quality, particularly as clinical decision support systems require transparent, interpretable justifications for medical decisions. Emerging models such as OpenAI's GPT-o series[10], Gemini 2.5 Pro[2,11], DeepSeek-R1[12], and Claude 4[13] have shown improved abilities to generate clinically coherent and valid reasoning. For instance, Brodeur et al. demonstrated that a GPT-4-based system achieved superhuman diagnostic and management performance across five clinical reasoning tasks, outperforming attending and resident physicians in accuracy, clarity of reasoning, and inclusion of high-risk ("cannot-miss") diagnoses[14]. Tordjman et al. benchmarked DeepSeek-R1 against GPT-o1 and Llama-3.1, demonstrating competitive performance across diagnostic tasks, including USMLE questions, NEJM challenges, and RECIST tumor classification, which highlights DeepSeek's strength in clinical reasoning[15].

Despite these promising developments, a critical and underexplored aspect of LLM-based clinical decision-making is causal reasoning—the ability to understand and infer cause-and-effect relationships. In medicine, causal inference is essential for tasks ranging from identifying disease etiology to evaluating treatment outcomes and recommending interventions[16,17]. As LLMs are increasingly integrated into high-stakes healthcare environments, evaluating and enhancing their capacity for causal reasoning is both urgent and necessary. Causal reasoning is fundamental to human intelligence, enabling us to understand the complex interplay of factors involved in clinical decision-making. Clinicians routinely assess symptoms, laboratory results, and comorbidities to determine the best course of action—decisions that rely on understanding cause-and-effect relationships rather than solely associations. While GenAI is widely expected to accelerate scientific discovery and improve medicine, its role in strengthening causal evidence in healthcare remains uncertain.

General frameworks for evaluating causal reasoning in LLMs have begun to emerge, many of which are grounded in Judea Pearl's Ladder of Causation[18], which outlines three levels of causal reasoning: association (rung 1), intervention (rung 2), and counterfactual (rung 3). Association refers to identifying statistical correlations without inferring causality. For example, one might observe that older, obese smokers tend to have higher HbA1c levels compared to younger, healthy individuals. However, this observation alone cannot tell us whether smoking or obesity causes elevated HbA1c as other unmeasured variables could be at play, and the direction of causality remains unclear. The second level, intervention, involves deliberately changing one variable to observe its impact on another. For instance, we might evaluate how quitting smoking affects HbA1c levels, enabling a more causal interpretation of the relationship. Finally, the third level, counterfactual, goes a step further by asking what would have happened under different conditions. For example, given that an older, obese smoker has an elevated HbA1c level, one might ask: "*Would their HbA1c level have been lower if they had never smoked or maintained a healthy weight earlier in life?*". This reasoning involves imagining alternate life scenarios and requires considering unmeasured confounders such as diet, physical activity, or genetic predisposition. These levels provide a structured foundation for assessing how well LLMs can understand and reason about causality in medical contexts.

Several foundational works have proposed benchmarks and frameworks to evaluate LLMs' causal reasoning capabilities. Jin et al. developed CLADDER, a 10,000-question dataset aligned with Pearl's Ladder of Causation, translating symbolic causal queries into natural language to assess LLMs' formal causal inference[19]. Similarly, Zhou et al. introduced CausalBench[20], a real-world benchmark evaluating LLMs' abilities across associational, interventional, and counterfactual, while Chen et al. presented CaLM (Causal evaluation of Language Models), a framework offering systematic performance metrics to measure reasoning depth in LLMs[21]. Complementing these, Khatibi et al. proposed ALCM (Autonomous LLM-Augmented Causal Discovery Framework), which integrates LLMs with causal discovery pipelines to generate interpretable causal graphs from observational data[22]. Vashishtha et al. introduced causal order as a more robust and accurate alternative to pairwise causal graphs when using expert feedback (including LLMs), proposing a novel triplet querying method that improves order inference accuracy, reduces cycles, and enhances performance across models and human annotators[23]. Kıcıman et al. conducted a behavioral study of LLMs, demonstrating their strong capabilities in generating causal arguments across diverse tasks, which often outperform traditional methods, while also highlighting their limitations and potential for integration with formal causal techniques[24]. Due to the importance of causality in medicine, Dang et al. introduced the Causal Roadmap, a rigorous framework for generating high-quality real-world evidence (RWE) from real-world data (RWD), emphasizing causal estimand specification, identifiability, and sensitivity analyses[25]. Building on this, Petersen et al. developed a causal copilot, an AI-guided tool using LLMs to assist researchers throughout the causal inference process, from formulating clinical questions to executing statistical plans, thus enhancing reproducibility and transparency in healthcare research[16]. Current AI models, including LLMs, are not well equipped to handle confounding, selection bias, and other challenges inherent in real-world data. Moreover, both causal claims derived from RWD (e.g., assessing whether elevated HbA1c levels causally increase the risk of developing type 2 diabetes) and predictions from black-box AI models (e.g., estimating cancer risk based on smoking history and lab test patterns without revealing which features drive the prediction) are often difficult to validate, an essential requirement for high-stakes clinical decision-making. Generating reliable causal evidence requires translating a clinical question into a well-defined "causal estimand,"[25] a process that is frequently hindered by poorly defined target populations, mismatched baseline time points, and vague or unrealistic treatment interventions.

This study systematically evaluates the causal reasoning abilities of LLMs in lab test-specific contexts, focusing on their performance across the three levels of Pearl's Ladder of Causation: association, intervention, and counterfactual. In alignment with the need to define precise causal estimands, we developed questions based on common laboratory tests such as Hemoglobin A1c (HbA1c), Creatinine, and Vitamin D, and paired them with clinically relevant causal factors, including age, gender, obesity, smoking, physical activity, medication use, and inflammation. To assess model performance, each causal question was verbalized using a structured prompt designed to elicit clear, medically accurate responses from two representative LLMs. The responses generated were evaluated by medically trained human experts, who rated both the answer and the reliability of reasoning.

**Results**

This study systematically evaluated the causal reasoning abilities of LLMs in the context of laboratory testing. Specifically, we assessed their performance across the three levels of Pearl's Ladder of Causation: association, intervention, and counterfactual reasoning. To ensure precise causal estimands, we generated 99 causal questions anchored in eight widely used blood tests: Hemoglobin A1c (HbA1c), Creatinine, Vitamin D, C-reactive protein (CRP), Cortisol, Low-Density Lipoprotein (LDL), High-Density Lipoprotein (HDL), and Albumin. For each test, we identified clinically relevant causal factors—including age, gender, obesity, smoking, physical activity, medication use, and inflammation—using authoritative medical sources such as MedlinePlus and the Mayo Clinic. The questions were developed within carefully defined clinical contexts that specified the target population, baseline characteristics, and potential interventions, allowing each to meaningfully support causal inference. To test model performance, the causal questions were verbalized into structured prompts designed to elicit accurate and clinically grounded responses. These were administered to two representative LLMs: GPT-o1 and Llama 3.2-8b-instruct. A detailed description of the causal question generation process is provided in the Methods section. Figure 1 summarizes the framework for evaluating LLM reasoning across three causal rungs: associational (is smoking associated with elevated HbA1c in older females?), interventional (what will happen to HbA1c if a patient quits smoking?), and counterfactual (was the smoking the cause of increased HbA1c?).

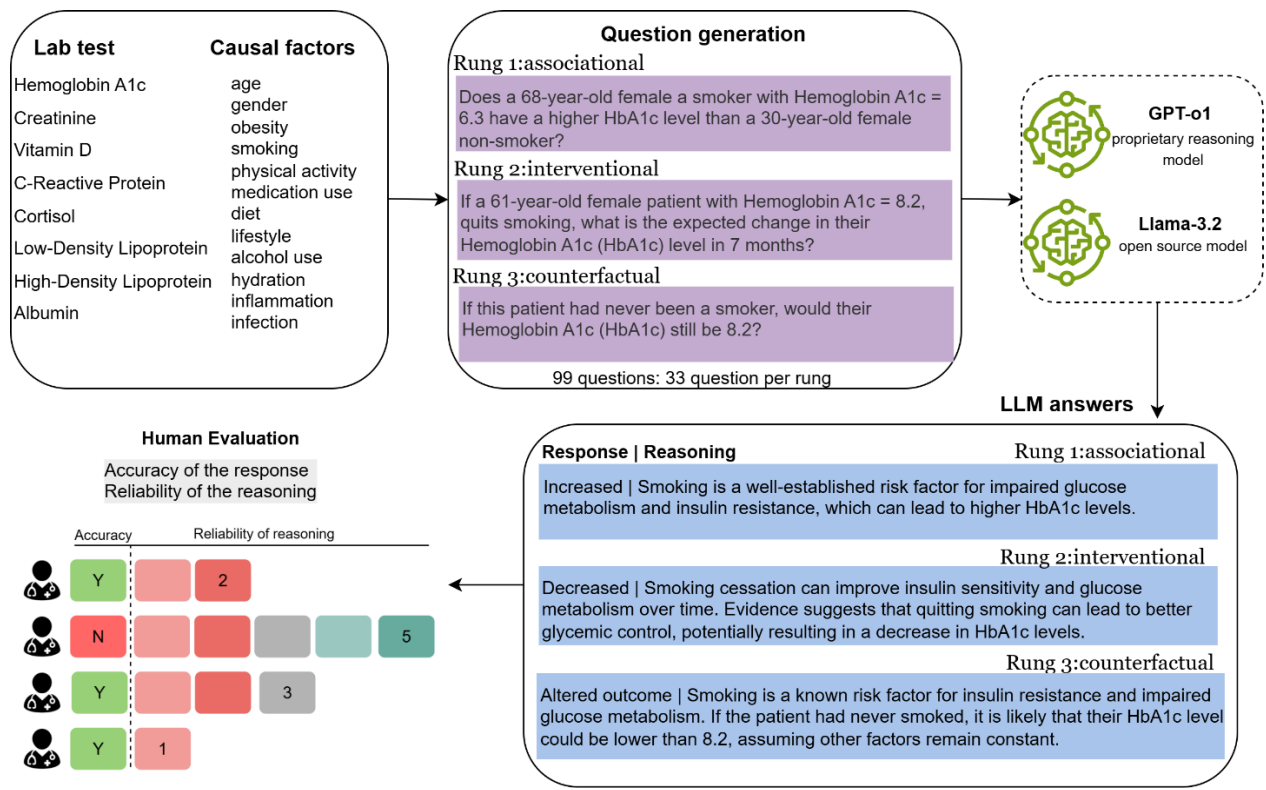

**Figure 1. Overview of the Pipeline for Generating and Evaluating Causal Questions and Reasoning.** The figure illustrates the end-to-end process used to generate causal questions based on laboratory tests and clinically relevant factors, produce reasoning responses using large language models, and evaluate these responses through expert review and quantitative metrics.

A total of 99 lab test-related causal questions were generated using Python code, with 33 questions for each of the three rungs of the causation ladder. LLM responses were expected to include a definitive answer indicating the direction or presence of a causal effect (e.g., *"Yes"*, *"No"*, *"Increased"*, *"Decreased"*, or *"Altered outcome"*), along with concise reasoning grounded in relevant risk factors, established clinical guidelines, or pathophysiological principles. The agreement rates varied across reasoning rungs and outcome types, highlighting differences in consistency. In rung 1: associational, perfect agreement was observed for "No" outcomes (1.00, n=9), while "Decreased" showed no agreement (0.00, n=5). Moderate agreement was found for "Increased" (0.56, n=9) and "Yes" (0.60, n=10). In rung 2: interventional, agreement was high for "Decreased" (0.91, n=22) and "Increased" (0.86, n=7), moderate for "No significant change" (0.50, n=2), but absent for both "Yes" and "Altered outcome" (0.00, n=1 each). In rung 3: counterfactual, agreement was very low for "Altered outcome" (0.04, n=24), yet high for "Yes"

(0.75, n=8) and perfect for "No" (1.00, n=1), suggesting stronger consensus when outcomes were binary. Overall, outcomes with clearer interpretations (e.g., "No" or "Decreased") tended to yield higher agreement, while more ambiguous or nuanced outcomes like "Altered outcome" introduced greater variability in judgment.

The LLMs response was independently assessed by four clinical raters, who classified the correctness of the answer using binary labels: "Yes" or "No". For the Association rung, 45 out of 66 responses (from both LLMs) had "Yes" as the majority response with an average agreement of 0.87, while eight responses had "No" (0.75), and 13 resulted in ties (0.50). The Intervention rung showed the highest agreement overall, with "Yes" as the majority in 60 of the 66 responses (0.95 agreement). The Counterfactual rung yielded 54 "Yes" responses (0.91), three "No" (0.75), and nine ties. When disaggregated by LLM, GPT-o1 consistently outperformed Llama 3.2 in agreement across all rungs, with particularly strong performance in the Counterfactual (0.92 vs. 0.77) and Association (0.81 vs. 0.76) categories. Both models performed well on Intervention questions, though GPT-o1 again had slightly higher rater consensus (0.92 vs. 0.90). This indicates that GPT-o1's outputs were generally more consistent with expert consensus, especially for complex causal reasoning tasks. The examples in Table 1 illustrate how LLMs responded to lab test-related causal questions using a context-rich prompt. These scenarios are grouped according to the three levels of Pearl's causal hierarchy—Rung 1 (Association), Rung 2 (Intervention), and Rung 3 (Counterfactual)—to showcase the diversity and complexity of causal reasoning tasks evaluated.

**Table 1.** Example use case scenarios generated by LLMs using the context-rich prompt, categorized by Causal Rung 1 (Association), Rung 2 (Intervention), and Rung 3 (Counterfactual).

| Question [Rung 1: Association]: Does a 68-year-old female a smoker with Hemoglobin A1c (HbA1c) = 6.3 have a higher Hemoglobin A1c (HbA1c) level than a 30-year-old female non-smoker? ||
|---|---|
| GPT-o1<br><br>*Response:* Increased<br><br>*Reasoning:* Smoking is a well-established risk factor for impaired glucose metabolism and insulin resistance, which can lead to higher HbA1c levels. Additionally, older age is associated with an increased risk of type 2 diabetes and higher HbA1c levels compared to younger individuals. | 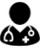 |
| Llama 3.2 | |
| *Response:* Yes<br>*Reasoning:* Elevated HbA1c levels are associated with impaired glucose metabolism, which can be influenced by aging, smoking, and other risk factors. The patient's age and smoking history increase her risk of insulin resistance, a key driver of elevated HbA1c levels. | Accuracy / Reliability of reasoning: N, 3; Y, 2; Y, 1; N, 3 |
| Question [Rung 2: Intervention]: If a 61-year-old female patient with Hemoglobin A1c (HbA1c) = 8.2, quits smoking, what is the expected change in their Hemoglobin A1c (HbA1c) level in 7 months? | |
| GPT-o1<br>*Response:* Decreased<br>*Reasoning:* Smoking cessation can improve insulin sensitivity and glucose metabolism over time. Evidence suggests that quitting smoking can lead to better glycemic control, potentially resulting in a decrease in HbA1c levels. Over a period of 7 months, it is reasonable to expect some improvement, although individual responses may vary. | Accuracy / Reliability of reasoning: Y, 1; Y, 3; Y, 2; Y, 1 |
| Llama 3.2<br>*Response:* Decrease<br>*Reasoning:* Quitting smoking improves insulin sensitivity and reduces inflammation, leading to better blood glucose control. This reduction in cardiovascular risk factors can help mitigate the impact of aging on insulin metabolism. | Accuracy / Reliability of reasoning: Y, 1; Y, 1; Y, 2; Y, 2 |
| Question [Rung 3: Counterfactual]: If this patient had never been a smoker, would their Hemoglobin A1c (HbA1c) still be 8.2? | |
| GPT-o1<br>*Response:* Altered outcome<br>*Reasoning:* Smoking is a known risk factor for insulin resistance and impaired glucose metabolism. If the patient had never smoked, it is likely that their HbA1c level could be lower than 8.2, assuming other factors remain constant. | Accuracy / Reliability of reasoning: Y, 1; Y, 3; Y, 3; Y, 2 |
| Llama 3.2 | |

*Response:* Decrease

*Reasoning:* The patient's current HbA1c level of 8.2 suggests that she has developed insulin resistance or impaired glucose metabolism over time, likely influenced by her smoking history. Without this exposure, it is reasonable to expect that her metabolic profile would be less compromised, leading to a lower HbA1c level.

Table 2 presents the average performance metrics (with standard deviation) for two large language models—GPT-o1 and Llama 3.2-8b-instruct—across binary classification and reasoning tasks stratified by causal rung: Associational, Interventional, and Counterfactual. For binary accuracy, GPT-o1 outperformed Llama 3.2 across all metrics and rungs, with an overall AUROC of 0.80 (±0.12) compared to Llama's 0.73 (±0.15). GPT-o1 also demonstrated higher precision (0.98 vs. 0.92), sensitivity (0.90 vs. 0.84), and specificity (0.70 vs. 0.62), particularly excelling in interventional and counterfactual reasoning scenarios where clinical implications are most critical. For the reasoning task, both models showed comparable AUROC values, with GPT-o1 scoring 0.76 (±0.13) and Llama 3.2 slightly higher at 0.81 (±0.01). However, GPT-o1 exhibited superior consistency in precision, sensitivity, and specificity, especially in the counterfactual category, where it achieved perfect performance (1.00 across all metrics). In contrast, Llama 3.2 displayed greater variability in specificity and sensitivity, particularly in counterfactual questions where its performance dropped (e.g., specificity = 0.50 ± 0.43). These results suggest that while Llama 3.2 can provide strong reasoning in certain cases, GPT-o1 delivers more stable and clinically aligned outputs across the spectrum of causal complexity.

| Metric | Model | Area under the receiver operating characteristic | | | | Precision | | | | Sensitivity | | | | Specificity | | | |
|---|---|---|---|---|---|---|---|---|---|---|---|---|---|---|---|---|---|
| | | Overall | Associational | Interventional | Counterfactual | Overall | Associational | Interventional | Counterfactual | Overall | Associational | Interventional | Counterfactual | Overall | Associational | Interventional | Counterfactual |
| Accuracy | gpt-o1 | 0.80 (0.12) | 0.75 (0.16) | 0.84 (0.24) | 0.84 (0.24) | 0.98 (0.01) | 0.96 (0.03) | 0.99 (0.02) | 0.99 (0.02) | 0.90 (0.08) | 0.82 (0.15) | 0.93 (0.06) | 0.93 (0.07) | 0.70 (0.26) | 0.67 (0.27) | 0.75 (0.50) | 0.75 (0.50) |
| | Llama-3.2-8b-instr | 0.73 (0.15) | 0.72 (0.19) | 0.70 (0.28) | 0.69 (0.20) | 0.92 (0.06) | 0.84 (0.12) | 0.98 (0.02) | 0.90 (0.09) | 0.84 (0.05) | 0.80 (0.15) | 0.92 (0.06) | 0.80 (0.10) | 0.62 (0.28) | 0.65 (0.24) | 0.50 (0.58) | 0.58 (0.40) |
| Reasoning | gpt-o1 | 0.76 (0.13) | 0.76 (0.13) | 0.78 (0.26) | 1.00 (0.00) | 0.97 (0.02) | 0.94 (0.03) | 0.96 (0.05) | 1.00 (0.00) | 0.96 (0.03) | 0.96 (0.03) | 0.97 (0.03) | 0.96 (0.06) | 0.57 (0.26) | 0.56 (0.24) | 0.58 (0.50) | 1.00 (0.00) |
| | Llama-3.2-8b-instr | 0.81 (0.01) | 0.81 (0.05) | 0.79 (0.35) | 0.78 (0.17) | 0.95 (0.03) | 0.91 (0.06) | 0.98 (0.02) | 0.95 (0.04) | 0.91 (0.08) | 0.88 (0.09) | 0.93 (0.12) | 0.92 (0.06) | 0.55 (0.37) | 0.58 (0.40) | 0.50 (0.58) | 0.50 (0.43) |

SFigure 1 shows the heatmap of binary accuracy ratings (0 = inaccurate, 1 = accurate) provided by four expert raters for 99 causal reasoning questions, comparing GPT-o1 (left) and Llama 3.2 (right). Figure 2 shows the heatmap of Likert-scale reasoning ratings (1 = excellent, 5 = poor) for the same set of questions, stratified by causal rung (association, intervention, counterfactual). Rows represent individual questions and columns represent raters, with lighter colors indicating higher reasoning quality and darker colors indicating poorer reasoning.

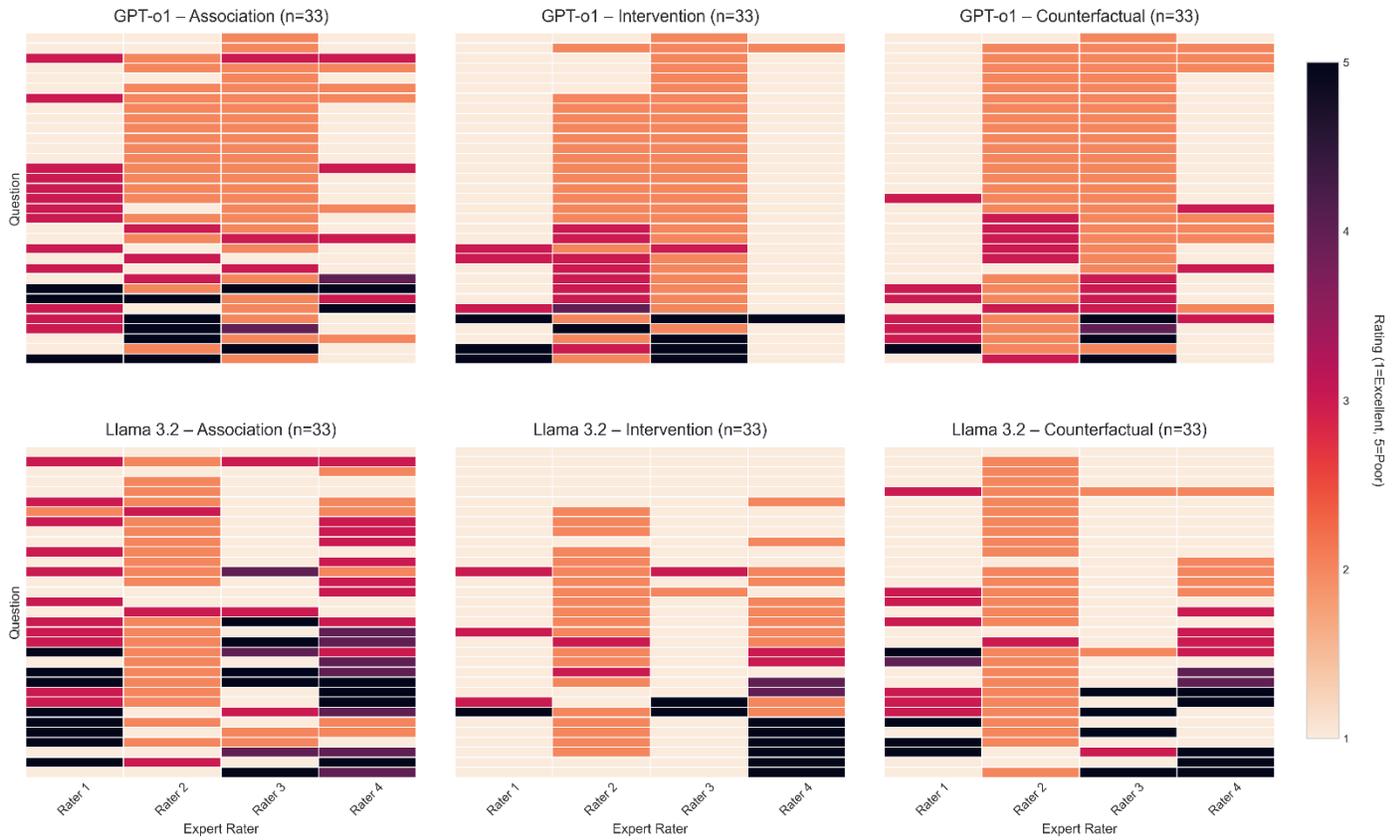

**Figure 2. Heatmaps of Reliable Reasoning Ratings (1 = Excellent, 5 = Poor) Stratified by Causal Rung for GPT-o1 and Llama 3.2.** Each heatmap panel shows expert ratings of reasoning quality for 33 causal reasoning questions per rung (Association, Intervention, Counterfactual) answered by two large language models (GPT-o1, top row; Llama 3.2, bottom row). Ratings were provided by four independent raters using a 5-point Likert scale (1 = high-quality reasoning, 5 = poor reasoning). Rows correspond to individual questions (Q1–Q99, evenly distributed across rungs), and columns to raters. Colors encode reasoning quality, with lighter tones representing better reasoning and darker tones representing poorer reasoning. Row order within each panel is sorted by inter-rater agreement to highlight consistency in judgments.

**Discussion**

This preliminary study introduced a structured framework to assess causal reasoning in LLMs (GPT-o1 and Llama-3.2_8b) using 99 clinically grounded laboratory test scenarios mapped to Pearl's Ladder of Causation—association, intervention, and counterfactual reasoning. Both models were able to generate medically relevant responses, with GPT-o1 consistently aligning more closely with expert consensus, particularly for complex counterfactual and associational tasks. Interventional questions, which often reflected widely known clinical relationships, saw strong performance from both models. Domain experts' ratings revealed that high-quality responses clearly explained causal pathways, reflected clinical mechanisms, and considered confounders such as age, gender, or comorbidities, whereas vague answers (e.g., "altered outcome") led to lower agreement and reliability scores. Quantitatively, binary outcomes that were inherently definitive ("No," "Decreased") produced the highest consensus (e.g., Association "No" = 1.00; Intervention "Decreased" = 0.91), while ambiguous counterfactual outcomes yielded near-zero agreement (0.04). GPT-o1 consistently outperformed Llama-3.2 across rungs, with the largest gaps on Counterfactual (0.92 vs. 0.77) and Association (0.81 vs. 0.76) questions, indicating more stable causal reasoning. These results underscore that while LLMs show potential for causal reasoning in medicine, their performance is uneven and often sensitive to question framing and clinical nuance. Our findings suggest that LLMs may be most reliable for well-established interventions but remain less stable for complex counterfactual reasoning. Future work should expand beyond lab tests to include history and physical exam findings, imaging, genomics, and longitudinal data, while integrating guardrails—such as retrieval augmentation, explicit assumption disclosure, and uncertainty tagging—to improve model reliability and trustworthiness in clinical settings.

We selected GPT-o1 and Llama-3.2-8b for causal reasoning because they embody two distinct yet complementary paradigms in current LLM development: a high-performing proprietary model and a leading small-scale open-source alternative with the potential for fine-tuning into causal reasoning–specific models. GPT-o1 delivers strong instruction-following and reasoning capabilities, while Llama-3.2-8b enables transparent, locally controlled experimentation, critical for ensuring reproducibility and safeguarding privacy in healthcare research. Compared to healthcare-specific models, these two offer broader generalization potential, allowing for unbiased evaluation of causal reasoning. We intentionally excluded medicine-specialized LLMs, as large electronic health record vendors continue to rely primarily on general-purpose commercial models in day-to-day clinical deployments.

GPT-o1's higher agreement suggests that larger or instruction-tuned proprietary models may internalize clinically coherent causal schemas better than smaller open-source counterparts. However, both models were strong in interventions, implying that widely available clinical knowledge (e.g., CRP levels fall after smoking cessation) is accessible even to smaller models. This pattern is evident in the heatmap (Figure 2), which is based on a 5-point Likert scale where 1 indicates a high rating and 5 indicates a poor one. GPT-o1's reasoning was rated closer to 1 (deep red) and more uniform, whereas Llama-3.2 showed broader dispersion, especially for counterfactual questions. Ties and mid-scale Likert scores cluster around questions with multiple plausible pathways or unclear temporal ordering. Grounding factors in authoritative sources (MedlinePlus, Mayo Clinic) likely reduced hallucinated mechanisms and gave raters a common frame of reference. Our few-shot plus instructions to reason step-by-step prompt probably improved coherence, yet did not eliminate failures on nuanced counterfactuals.

Analyzing example cases reveals a clear evaluation pattern: medically trained domain experts favored responses that not only stated the correct direction but also provided a well-reasoned explanation grounded in medical knowledge. In contrast, they rated responses lower when the reasoning was vague, inconsistent, or poorly worded—even if the final answer matched the consensus.

For instance, Llama-3.2's HbA1c intervention item (Q103: "*If a 63-year-old male with Hemoglobin A1c = 9.7 quits smoking, what is the expected change in HbA1c in 4 months?*") was unanimously rated (4/4 "Yes"; all Likert = 1) because the model explicitly articulated that smoking cessation attenuates insulin resistance and chronic low-grade inflammation—key drivers of sustained hyperglycemia—thereby lowering average glucose exposure and, consequently, HbA1c over the ensuing months. A similar pattern was observed for lab test LDL (Q196: "*If a 53-year-old male with elevated LDL quits smoking, what is the expected change in LDL?*"), where the model again earned perfect scores by linking cessation to reduced oxidative stress, improved endothelial function, and favorable lipid metabolism, logically predicting a decrease in LDL concentrations. GPT-o1 demonstrated comparable mechanistic fidelity on analogous prompts (e.g., "*Smoking cessation can improve insulin sensitivity and lower chronic inflammation, which reduces HbA1c*

*over time*" and "*Smoking cessation is associated with improved lipid metabolism and modest LDL reductions*"), although a minority of raters nudged Likert scores to 2, reflecting minor reservations about completeness or nuance rather than fundamental causal soundness.

In contrast, the incorrect responses help explain why lower reasoning scores (4–5) were often associated with disagreement among domain experts. In an Association CRP test question (Q111: "*Do 66-year-old male smokers have higher CRP than non-smokers?*"), Llama-3.2 answered "No" and offered only a vague assertion of "no clear difference," disregarding extensive evidence that tobacco exposure upregulates inflammatory pathways (e.g., NF-κB activation) and elevates acute-phase reactants such as CRP; three raters therefore judged the answer incorrect. Likewise, for HbA1c Association (Q105: "*Does a 54-year-old female smoker with HbA1c = 6.2 have a higher HbA1c than a 24-year-old female non-smoker?*"), the model's "No" aligned with the majority binary decision, but the explanation—hinging on age and "shorter duration of smoking" without interrogating dose–response relationships, adiposity, or medication status—was deemed superficial (Likert 5/3/1/5).

When answering questions about kidney function using creatinine levels, GPT-o1 consistently provided correct answers and clear reasoning across all three levels of causality. All four domain experts agreed with GPT-o1's answers on both the association and intervention prompts. Its association response explained that "*elderly males generally experience age-related changes in renal function, including reduced glomerular filtration rate (GFR), leading to higher creatinine levels compared to younger females,*" and was rated highly (Likert 1–2 by all four raters). In the intervention scenario, GPT-o1 reasoned that "*increasing water intake can improve renal perfusion and hydration, leading to better creatinine clearance,*" again earning full agreement and Likert scores between 1 and 2. For the counterfactual—if the male patient had been female—three domain experts agreed with GPT-o1's "Yes" response, but one disagreed. Reasoning was rated slightly lower (Likert 1–3), suggesting mild uncertainty about how sex alone would normalize values. In contrast, Llama-3.2 showed a more variable performance. While its intervention response was supported by all domain experts— "we can expect a decrease… a study found significant reductions in creatinine levels"—one domain expert rated the reasoning a 5 for being too reliant on external data without tightly connecting it to the patient. For the association and counterfactual questions, only half the domain experts agreed with Llama-3.2's answers, and reasoning scores ranged widely (Likert 1 to 5), with raters pointing to vagueness, internal contradiction, or overgeneralization about age and gender effects.

For lab test cholesterol (LDL) questions, GPT-o1 again performed best on intervention and counterfactual tasks. In the intervention prompt about quitting smoking, GPT-o1 stated, "*smoking does not have a direct, immediate effect on LDL… the benefits are seen in inflammation and vascular function,*" which all four domain experts endorsed, giving Likert scores between 1 and 3. The counterfactual reasoning—"*absence of smoking would likely result in a lower LDL level*"— was also accepted by all, with strong ratings (Likert 1–2). However, on the association prompt (comparing a 58-year-old smoker to a 29-year-old non-smoker), GPT-o1 answered "No" and failed to elaborate on smoking's known effects on LDL; only one domain expert agreed, and reasoning scores were poor (Likert 5/5/2/5). Llama-3.2, in contrast, offered much stronger explanations, such as "*smoking damages the endothelium, increases inflammation… contributing to increased LDL levels,*" which was clinically accurate responses. Three domain experts agreed with this association answer and rated the reasoning between 1 and 3. For the counterfactual, Llama-3.2 again gave a comprehensive explanation, referencing multiple studies, and three domain experts agreed with both the answer and reasoning (Likert 1–3), though one domain expert disagreed with the final label. Overall, domain expert ratings revealed that GPT-o1 was more consistent in aligning answers with clinical consensus, while Llama-3.2 stood out for its richer scientific explanations, though sometimes at the cost of clarity or alignment with expert judgment.

Collectively, these cases underscore a consistent evaluative heuristic: domain experts privileged responses that traced a logically consistent causal pathway (e.g., smoking cessation → reduced systemic inflammation/insulin resistance → lower HbA1c) and penalized those lacking mechanistic specificity (e.g., failing to explain how smoking influences lipid metabolism or how hydration affects renal function), internal coherence (e.g., contradicting the stated lab values or patient demographics), or appropriate consideration of clinical confounders and effect modifiers (e.g., overlooking age, sex, medication use, or comorbid conditions that could influence lab results). Notably, GPT-o1 never produced a continuous sequence of Likert 1 scores, with isolated 2–3 scores signaling the domain experts' stringent expectations for completeness, precision, and clinical plausibility.

The study has certain limitations; we limited ourselves to eight common blood tests; generalizability to imaging, genomics, or longitudinal EHR trajectories remains unknown. Results cannot be extrapolated across the rapidly evolving LLM landscape. Collapsing rich causal statements into a small set of categorical answers ("Altered outcome") may have inflated ambiguity. Our findings caution against assuming uniform "causal competence" in LLMs. Models are reliable for well-trodden interventional knowledge but unstable for counterfactual reasoning that requires disentangling multiple interacting causes. For deployment, LLM guardrails such as retrieval-augmented generation (RAG) with clinical causal guideline documents, explicit requirement to state assumptions, and automatic uncertainty tagging could mitigate risk. Evaluations should move beyond surface correctness to probe whether models identify appropriate confounders, temporal ordering, and effect modifiers.

**Methods**

**Study Design**

This study systematically evaluates the causal reasoning abilities of LLMs in lab test-specific contexts, focusing on their performance across the three levels of Pearl's Ladder of Causation: association, intervention, and counterfactual. In alignment with the need to define precise causal estimands, we generated the questions grounded in eight common blood laboratory tests: Hemoglobin A1c (HbA1c), Creatinine, Vitamin D, C-reactive protein (CRP), Cortisol, Low-Density Lipoprotein (LDL), High-Density Lipoprotein (HDL), and Albumin. For each test, we identified clinically relevant causal factors—such as age, gender, obesity, smoking, physical activity, medication use, and inflammation—by using reputable medical sources like MedlinePlus and the Mayo Clinic. This process involved clearly specifying the clinical context, target population, baseline characteristics, and potential interventions to ensure that each question could meaningfully support causal inference. Questions were then crafted to reflect the three causal levels: Associational (e.g., *comparing HbA1c levels between older obese smokers and younger healthy individuals*), Interventional (e.g., *evaluating the impact of quitting smoking on HbA1c levels*), and Counterfactual (e.g., *would HbA1c levels have been lower if the patient were not obese?*). To assess model performance, each causal question was verbalized using a structured prompt designed to elicit clear, medically accurate responses from two representative LLMs: GPT-o1(a widely used proprietary reasoning model) and Llama-3.2(a popular open-source model). The responses generated were evaluated by medically trained human experts, who rated both the answer (Binary [yes/no]) and the clinical validity of the reasoning (Likert scale [1-5]). GPT-o1 (*o1-2024-12-17*) was accessed via Application Programming Interfaces (API) calls to OpenAI, while Llama-3.2(*8b_instruct*) was downloaded locally and accessed using an open-source application called Jan (https://jan.ai/). For both models, the temperature was set to 0 and the top-p parameter to 1.

For causal question answering and reasoning, we employed a few-shot learning approach, where the model was provided with multiple example outputs formatted in a tab-separated structure. The prompt (detailed in Box 1) explicitly instructed the model to generate responses where each answer was paired with its corresponding reasoning. The model was tasked with analyzing clinical scenarios and applying medical knowledge to the aforementioned laboratory markers, systematically explaining causal relationships through the three rungs of reasoning. Specific instructions were provided that each question must be answered using step-by-step reasoning grounded in clinical evidence, physiological mechanisms, and standard medical guidelines. For rung-1 association questions, the prompt instructed the LLM to identify statistical or observational relationships between variables and detect trends and risk factors. For rung-2 intervention questions, the model assessed how changes in risk factors, e.g., lifestyle modifications or medications, would impact lab markers and clinical outcomes. When addressing rung-3 counterfactual questions, the LLM "imagined" how alterations in a patient's history or behavior would have changed the outcome, or more specifically, if it was a specific factor that led to their current status.

Table 3 presents a summary of the unique causal factors used in generating causal reasoning questions for each laboratory test. For each test (e.g., Hemoglobin A1c, Creatinine, Cortisol), the associated causal factors include demographic (e.g., age, gender), lifestyle (e.g., smoking, physical activity), clinical (e.g., diabetes, kidney disease), and treatment-related variables (e.g., medication use). These factors reflect known physiological and clinical influences on the respective lab test values and were used to construct association, intervention, and counterfactual reasoning questions.

**Table 3. Causal Factors Associated with Each Laboratory Test**

| Lab Test | Causal Factors |
|---|---|
| Albumin | hydration, infection, inflammation, kidney disease, liver disease, medication use |
| C-Reactive Protein (CRP) | autoimmune diseases, cardiovascular disease, medication use, infection, obesity, smoking |
| Cortisol | age, chronic illness, gender, obesity, medication use |
| Creatinine | age, diabetes, gender, high protein intake, hydration status, kidney disease |
| Hemoglobin A1c (HbA1c) | age, gender, medication use, physical activity, smoking |
| High-Density Lipoprotein (HDL) | alcohol consumption, medication use, physical activity, smoking |
| Low-Density Lipoprotein (LDL) | diabetes, lack of exercise, medication use, obesity, smoking |
| Vitamin D (25-hydroxyvitamin D) | age, kidney disease, gender, lifestyle, medication use |

The selected panel of eight laboratory tests—Albumin, C-Reactive Protein (CRP), Cortisol, Creatinine, Hemoglobin A1c (HbA1c), High-Density Lipoprotein (HDL), Low-Density Lipoprotein (LDL), and Vitamin D (25-hydroxyvitamin D)—represents a clinically diverse set of biomarkers commonly used in diagnosing and monitoring chronic diseases, metabolic status, organ function, and systemic inflammation. Each lab test is influenced by a distinct set of causal factors that modulate its levels and clinical interpretation. For instance, Albumin, a key marker of nutritional status and liver/kidney function, is sensitive to hydration, infection, inflammation, and organ dysfunction. CRP, an acute-phase reactant, reflects systemic inflammation and is elevated in the context of autoimmune disorders, cardiovascular disease, obesity, and smoking. Cortisol, the primary stress hormone, is affected by age, gender, chronic illness, and medication, serving as a window into endocrine and psychological health. Creatinine, a renal function marker, varies with age, hydration status, protein intake, and kidney disease. HbA1c, a marker of long-term glucose control, is influenced by age, lifestyle, and medication use, making it central to diabetes management. HDL and LDL, critical indicators of lipid metabolism and cardiovascular risk, are shaped by smoking, physical activity, alcohol consumption, and metabolic disorders such as obesity and diabetes. Lastly, Vitamin D status is governed by age, ethnicity, gender, lifestyle, and medication, with implications for bone health, immunity, and chronic disease prevention.

Understanding how these diverse causal factors interact with each lab test enhances the precision of clinical assessments and supports the development of context-aware LLM reasoning models.

---

**Box 1. Few-shot prompt used in this study to generate answers and reasoning for causal questions.**

```
Please act as a medical professional specializing in laboratory diagnostics and
clinical decision-making. Your task is to analyze causal relationships related
to various laboratory markers (e.g., HbA1c, creatinine, CRP, cortisol) by
assessing association, intervention effects, and counterfactual scenarios.

You must approach each question by:

Analyzing the Scenario: Carefully examine the provided medical context.
Applying Medical Knowledge: Utilize established research, clinical guidelines,
and pathophysiological principles.
Reasoning Step-by-Step: Break down causal relationships systematically.
Providing Evidence-Based Answers: Support responses with medical literature,
standard clinical guidelines, or physiological mechanisms.

The question-and-answer format should be:
Association Q: [Question comparing two groups or conditions] A:
[Yes/No/Increased/Decreased/etc.] Reasoning: [Brief explanation based on
relevant risk factors, pathophysiology, or established evidence]

Intervention Q: [Question describing an intervention and its likely effect on a
lab marker or clinical outcome] A: [Expected direction of change or result]
Reasoning: [Explanation of how the intervention mechanistically alters the lab
marker or outcome, supported by evidence or guidelines]

Counterfactual Q: [Hypothetical scenario where a key risk factor or condition is
absent or changed] A: [Yes/No/Altered outcome] Reasoning: [Explanation of how
the absence or alteration of the factor would affect the outcome, based on
pathophysiology or epidemiological data]

Please produce the response and reasoning in tab-separated format '\t'

Example output
Response \t Reasoning
Yes     Older age, obesity, and smoking are all associated with insulin resistance
and impaired glucose metabolism, leading to elevated HbA1c levels.\n
Decrease     Regular exercise improves insulin sensitivity and enhances
peripheral glucose uptake, gradually lowering blood glucose levels and thus
HbA1c over time.\n
No     Obesity is a key driver of insulin resistance. Without excess adiposity,
the patient's insulin sensitivity would likely be better, contributing to lower
HbA1c levels.\n
```

Four medically trained human experts were instructed to evaluate the performance of GPT-o1 and Llama-3.2_8b in response to a curated set of clinical causal reasoning questions. Each domain expert independently assessed the responses using a structured evaluation rubric. They judged accuracy based on whether the final answer was categorically correct (yes/no/NA) and rated the reliability of reasoning—the clarity, scientific validity, and clinical usefulness of the explanation—on a 5-point Likert scale (1 = highly reliable to 5 = highly unreliable). Participation was voluntary, with the option to complete the full set or only a portion, and to withdraw at any time. No compensation was offered. Domain experts were asked to return their evaluations within ten days, after which two weekly reminders would be sent before the survey closed.

Inter-rater agreement on LLM evaluations was assessed using two complementary approaches: Fleiss' kappa with Conger's exact procedure for binary correctness judgments and Kendall's coefficient of concordance (W) with ties correction for ordinal reasoning quality ratings. As shown in SFigure 2, binary correctness (Yes/No) ratings across four domain expert raters revealed higher agreement for intervention-type questions—particularly for GPT-o1—while counterfactual scenarios, especially for LLaMA 3.2, showed the lowest agreement. SFigure 3 illustrates ordinal concordance on Likert-scale reasoning quality ratings (1 = highly reliable, 5 = poor), where agreement again peaked for intervention tasks and was weakest for counterfactuals, indicating greater subjectivity in hypothetical reasoning. Finally, SFigure 4 highlights domain expert-specific variability in median reliability ratings, underscoring differences in subjective perceptions of model performance across raters. To evaluate the consistency and reliability of model performance, we computed a comprehensive set of binary classification metrics, including Area Under the Receiver Operating Characteristic Curve (AUROC), Precision, Sensitivity, and Specificity. Analyses were stratified by the rung level of causality. Each domain expert's binary or Likert-based evaluation was compared against a consensus ground truth—defined as a majority vote for binary responses or the median rating for Likert-scale assessments. All analyses were conducted using R (v4.3.0), with packages including DescTools, grDevices, irr, ROCR, stringr, and xlsx. Python packages such as matplotlib, seaborn, pandas, and numpy were used to generate visualizations. Data and reproducible code are provided in the supplementary materials.

**Code Availability**

All code, inference outputs, and scripts used to generate the results in this study are publicly accessible at https://github.com/balubhasuran/LLM_Causality_LabTest . Experiments were performed using Python 3.11 with the following open-source libraries: matplotlib =3.8.4, numpy =1.24.4, openai =1.97.1, pandas =1.5.3, scikit_learn =1.3.2, scipy =1.11.4, seaborn =0.13.2, statsmodels = 0.14.0. Statistical analyses were conducted in R 4.5.1 using the following packages: tidyverse =v1.3.0, ggplot2 = v3.5.2, lme4 = v1.1-37, DescTools = v0.99.60, grDevices = base, irr = v0.84.1, ROCR = v1.0-11, stringr = v1.5.1, and xlsx = v0.6.5.

**Acknowledgment**

This work was supported by the Agency for Healthcare Research and Quality grant R21HS029969 (PI: Z.H.). This project was also partially supported by the University of Florida-Florida State University Clinical and Translational Science Award, which is supported in part by the National Institutes of Health (NIH) National Center for Advancing Translational Sciences under award UL1TR001427. We would like to thank Dr. Julian R. Mark (Department of Neuroscience, University of Florida, College of Medicine) for evaluating the LLM-generated responses.